\documentclass[letterpaper, 10 pt, conference]{ieeeconf}  % Comment this line out if you need a4paper

\IEEEoverridecommandlockouts                              % This command is only needed if 
\usepackage{times} % assumes new font selection scheme installed
\usepackage{multirow}
\usepackage{amsmath,amssymb,amsfonts}
\usepackage{xcolor}
\usepackage{siunitx}
\usepackage{booktabs}
\usepackage{subcaption}
\usepackage[hidelinks]{hyperref}
\usepackage{flushend}

\usepackage[
    detect-all=true,
    detect-family=true,
    group-digits=false,
    separate-uncertainty=true
]{siunitx}
\usepackage{pgfplots}
\pgfplotsset{compat=newest}
\usepgfplotslibrary{groupplots}
\usepgfplotslibrary{dateplot}
\usetikzlibrary{arrows.meta}
\usetikzlibrary{backgrounds}
\usepgfplotslibrary{patchplots}
\usepgfplotslibrary{fillbetween}
\pgfplotsset{
     layers/standard/.define layer set={
         background,axis background,axis grid,axis ticks,axis lines,axis tick labels,pre main,main,axis descriptions,axis foreground
     }{
         grid style={/pgfplots/on layer=axis grid},
         tick style={/pgfplots/on layer=axis ticks},
         axis line style={/pgfplots/on layer=axis lines},
         label style={/pgfplots/on layer=axis descriptions},
         legend style={/pgfplots/on layer=axis descriptions},
         title style={/pgfplots/on layer=axis descriptions},
         colorbar style={/pgfplots/on layer=axis descriptions},
         ticklabel style={/pgfplots/on layer=axis tick labels},
         axis background@ style={/pgfplots/on layer=axis background},
         3d box foreground style={/pgfplots/on layer=axis foreground}
     },
 }

\usepackage{graphicx}
\usepackage{url}
\usepackage[capitalize]{cleveref}
\usepackage{csquotes}
\usepackage{vector}

\usepackage{algorithm}
\usepackage[noend]{algpseudocode}

\usepackage{epsfig}
\usepackage{mathptmx}

\newcommand{\ra}[1]{\renewcommand{\arraystretch}{#1}}
\newcolumntype{R}{>{$}r<{$}}

\newcommand{\param}{\theta}
\newcommand{\hor}{\tau}

\def\baselinestretch{1}
\AtBeginDocument{
}

\title{\LARGE \bf
Model Identification Adaptive Control with $\rho$-POMDP Planning
}
\author{Michelle Ho,$^1$ Arec Jamgochian,$^{1,2}$ and Mykel J. Kochenderfer$^1$%  <-this % stops a space
\thanks{$^{1}$Stanford University, Stanford, CA 94305 USA
{\tt\small \{mtho, arec, mykel\}@stanford.edu}}%
\thanks{$^{2}$TerraAI, Redwood City, CA 94063 USA}%
\thanks{*Toyota Research Institute provided funds to support this work.}% <-this % stops a space
} 

\begin{document}

\maketitle
\begin{abstract}
Accurate system modeling is crucial for safe, effective control, as misidentification can lead to accumulated errors, especially under partial observability. We address this problem by formulating informative input design and model identification adaptive control (MIAC) as belief space planning problems, modeled as partially observable Markov decision processes with belief-dependent rewards ($\rho$-POMDPs). We treat system parameters as hidden state variables that must be localized while simultaneously controlling the system. We solve this problem with an adapted belief-space iterative Linear Quadratic Regulator (BiLQR). We demonstrate it on fully and partially observable tasks for cart-pole and steady aircraft flight domains. Our method outperforms baselines such as regression, filtering, and local optimal control methods, even under instantaneous disturbances to system parameters.
\end{abstract}

\section{Introduction}

Accurate system dynamics models are essential for safe, effective control since incomplete knowledge or simplifying assumptions can lead to misspecifications. System identification uses observed data to estimate model parameters for better predictions, fault detection, and robust control \cite{ljung1999system}. For example, an aircraft approximated as a linear model can be rendered inaccurate due to continuous wind disturbances, but its parameters can be updated online to compensate~\cite{ott2024informative}. This approach is vital in many fields, including robotics~\cite{johansson2000statespace}, biomedical engineering \cite{wiens2015novel}, and finance \cite{los2006system}.

Model identification adaptive control (MIAC) extends system identification by simultaneously learning system parameters and controlling the system to complete a desired objective \cite{oreg2019model}.
Though its online parameter estimation allows it to adapt to external disturbances and time-varying dynamics, typical MIAC methods assume full observability \cite{nishimura2021rat, wahlberg_optimal_2010, slade2017simultaneous}, which is problematic in settings with limited state information. Moreover, their extension to continuous state and action spaces can introduce intense computational effort. 
% \begin{figure}[H]
%     \centering
%     \includegraphics[width=0.8\linewidth]{figs/ilqr_types.jpeg}
%     \caption{Comparison of iLQG, and BiLQR: iLQG uses the mean of the Gaussian filter belief for planning. In contrast, belief-state iLQR (BiLQR) plans over the full Gaussian belief-state, enabling quadratic regulation of the terminal belief covariance besides controlling the mean. We 
%     perform MIAC using BiLQR to take actions that 
%     use BiLQR to simultaneously control a system while minimizing uncertainty over system parameters.}
%     \label{fig:ilqr_schemes}
% \end{figure}

% To enhance parameter estimation, we use informative input design, which selects control inputs that maximize information gain about the parameters \cite{ott2024informative}. We formulate MIAC as a partially observable Markov decision process with belief-dependent rewards ($\rho$-POMDP) \cite{araya2010pomdp} that balances control performance with system identification. Here, a belief state represents a probability distribution over the true system state and parameters, capturing both our estimates and their uncertainties. Since belief space dynamics can be nonlinear, stochastic, and underactuated, we efficiently solve the $\rho$-POMDP using an adapted belief-state iterative Linear Quadratic Regulator (BiLQR) that plans over Gaussian belief states and penalizes state uncertainty \cite{platt2010belief}. This combined approach allows us to identify system parameters more quickly by prioritizing information-gathering actions while maintaining safe control through robust state estimation.

We address planning in partially observable environments with continuous state and action spaces. We treat system parameters as hidden state variables, which we estimate with informative input design, which selects controls that maximize information gain \cite{ott2024informative}. We reformulate system identification under informative input design and MIAC as partially observable Markov decision processes with belief-state dependent rewards ($\rho$-POMDPs \cite{araya2010pomdp}), where belief-states include both the mean and uncertainty of true states. Since belief-state dynamics are nonlinear, stochastic, and underactuated, we efficiently solve the $\rho$-POMDP using an adapted belief-state iterative Linear Quadratic Regulator (BiLQR) that plans over Gaussian beliefs and penalizes uncertainty \cite{platt2010belief}. This approach quickly identifies system parameters by prioritizing information-gathering actions while maintaining effective control through robust state estimation.

% In this paper, we address the challenge of planning in partially observable environments with continuous action spaces. We treat system identification and MIAC as planning problems in a partially observable environment with system parameters as additional hidden states. We formulate MIAC as a partially observable Markov decision process with belief-dependent rewards ($\rho$-POMDP) \cite{araya2010pomdp} that balances control with system identification. Since belief space dynamics may be nonlinear, stochastic, and underactuated, we solve this $\rho$-POMDP efficiently with an adapted version of belief-state iterative Linear Quadratic Regulator (BiLQR). BiLQR plans over Gaussian belief states, allowing for objectives that penalize state uncertainty \cite{platt2010belief}. We show that planning with beliefs over system parameters in this fashion allows us to identify system parameters more quickly by prioritizing information-gathering actions while maintaining safe control.

We present our approach on both fully and partially observable problems for a cart-pole balancing example and a steady aircraft flight example. In these problems, BiLQR outperforms common approaches that combine filtering or regression for informative input design and model predictive control approaches for MIAC. BiLQR not only achieves control objectives while identifying system parameters more quickly and accurately, but also adapts to time-varying dynamics. To our knowledge, this is the first work to formulate MIAC as a $\rho$-POMDP jointly over system states and parameters, providing efficient solutions to MIAC problems while reasoning over uncertainty about system parameters. 

In summary, our contributions include:
\begin{itemize}
\item formulating system identification and MIAC as a $\rho$-POMDP over system states and parameters,
\item solving the $\rho$-POMDP using BiLQR more accurately than the baselines, and 
\item demonstrating our approach in fully and partially observable settings, and with time-varying dynamics.
\end{itemize}

In this paper, we first provide an overview of system identification, MIAC, and POMDPs. We then discuss our approach for formulating MIAC problems as $\rho$-POMDPs and solving them with BiLQR. Finally, we demonstrate our approach against various baselines in fully and partially observable environments with unknown system parameters.

\section{Background}
\subsection{System Identification and Adaptive Control}
System identification uses observed data to learn parameters $\param$ that dictate a dynamics model $f(s,a)$ \cite{ljung1999system}. Methods for fully observable systems include regression \cite{kopp1963linear}, parametric modeling \cite{gedon2021deep}, and informative input design, which selects controls to maximize information about model parameters \cite{ott2024informative}. Some filtering methods, like Kalman Filtering, can address partial observability with these methods \cite{valasek2003observation}. However, decoupling parameter and state estimation can lead to suboptimal information flow, limiting the accuracy of both parameter and state estimates. 

% IID optimizes input signals to maximize the quality of observed data, enhancing filtering performance and improving parameter estimation \cite{ott2024informative}.

Several methods also combine model identification and control. Many decouple system identification from control for quicker estimation, such as by using expectation maximization \cite{baum1970maximization}, planning for worst-case parameter estimation or distributions \cite{nishimura2021rat}, optimizing data generation for better estimation \cite{wahlberg_optimal_2010}, or sampling-based methods \cite{slade2017simultaneous}. Model reference adaptive control (MRAC) selects controls so the system follows a reference model that represents the desired behavior \cite{schreier2012modeling}. Model identification adaptive control simultaneously learns system parameters and controls the system~\cite{oreg2019model}. Compared to MRAC, MIAC’s online estimation allows flexible responses to disturbances and time-varying dynamics~\cite{oreg2019model}. Still, existing MIAC methods do not explicitly account for partial observability or consider actions that balance control and parameter uncertainty reduction simultaneously. 

\subsection{POMDPs}
A partially observable Markov decision process (POMDP) is a formulation for systems with observations that provide incomplete information about the state. A finite horizon POMDP is defined by the tuple $(\mathcal{S}, \mathcal{A}, \mathcal{O}, T, O, R, \hor)$ consisting of state, action, and observation spaces $\mathcal{S}, \mathcal{A}, \mathcal{O}$, a dynamics model $T$ mapping states and actions to a distribution over resulting states, an observation model $O$ mapping an underlying transition to a distribution over emitted observations, a reward function $R$ that maps the underlying transition to an incremental reward, and finite time horizon $\hor$. A policy $\pi$ generates actions from an initial state distribution $b_0$, a history of actions $a_{0:t}$, and observations $o_{1:t}$, which together can be represented concisely as an instantaneous belief distribution over states $b_t$ where $b_t(s)=p(s_t=s\mid b_0, a_{0:t}, o_{1:t})$. An optimal policy maximizes the expected cumulative reward over the time horizon: 
\begin{align}
\max_\pi & \ V^\pi(b_0)= \mathbb{E}_{T, O}\left[\sum_{t=0}^\hor R(b_t,a_t) \mid b_0 \right]  \label{eq:cpomdp-objective-reward} \text{,}
\end{align}
where the belief-based reward function returns the expected reward from transitions from states in the belief \cite{astrom1965optimal}.

Traditional POMDPs model state-based rewards and cannot explicitly model an agent's goal to reduce state uncertainty. To address this, $\rho$-POMDPs modify the standard POMDP framework by introducing belief-dependent rewards that favor actions that reduce the uncertainty (i.e., entropy) in the belief state \cite{araya2010pomdp}. They have been applied to active sensing problems to balance between strategically gathering information and completing tasks \cite{ott2023sequential}. This work can be seen as a dual control problem with belief-dependent rewards.

Exact planning in large POMDPs is generally intractable. While small, discrete POMDPs can be solved offline with approximate methods \cite{kurniawati2009sarsop}, larger ones require online, receding-horizon planning \cite{silver2010monte}. Sampling-based tree search extends to continuous state and action spaces \cite{slade2017simultaneous} but struggles in high-dimensional control problems. Trial-and-error techniques like reinforcement learning can learn effective policies in large problems, but can be very data inefficient \cite{deisenroth2011pilco}. Under certain assumptions, these problems can be addressed using optimal control methods like model predictive control (MPC) or Linear Quadratic Regulator (LQR) from a mean state estimate \cite{gonzalez2005robust}. These methods can be extended to plan over nonlinear dynamics with iterative linearization (e.g. iLQR), or to plan over full belief-state dynamics rather than planning from a mean (BiLQR) \cite{platt2010belief}. 

% Prior methods for belief-space planning like belief-space iLQR \cite{platt2010belief} and point-based solvers \cite{kurniawati2009sarsop} typically assume known dynamics. In contrast, we use BiLQR to plan directly in belief space, planning for state and parameter uncertainty propagation. We adopt BiLQR since it extends LQR, a standard optimal control method, to nonlinear, partially observable systems. By formulating MIAC as a $\rho$-POMDP, we enable the use of belief-space solvers like BiLQR. 

% Approximate Bayesian RL approaches such as PILCO \cite{deisenroth2011pilco} guide policy search by propagating uncertainty through Gaussian Process models with batch learning. 

% Notably, belief-state iLQR (BiLQR), an extension of iLQR that plans over future belief states using known dynamics was successfully applied to belief space planning by \citeauthor{platt2010belief}. This approach addresses nonlinear systems with partial observability \cite{platt2010belief}, but has not been applied to MIAC, where dynamics may be misidentified. 

\section{Methodology}

% \subsection{Preliminaries}

\subsection{Problem Formulation}
As stated previously, the goal of system identification is to learn the system parameters. The goal of \textit{informative input design} for system identification is to choose actions that decrease uncertainty over system parameters. Consider a prior distribution over possible system parameters
$p(\param)$ with shorthand $b_{\param}$. Let $b_{\param, \hor}$ be the shorthand for the posterior distribution over system parameters consistent with observations through the time horizon $\hor$. With an uncertainty measure $H$ of the belief (e.g., Shannon entropy), we can formulate the objective for active system identification as: 

\begin{equation}
    \min_\pi \ \mathbb{E}[H(b_{\param,\hor})] \text{.}
\end{equation}
That is, the optimal policy $\pi$ results in minimal expected uncertainty in the parameters at the final time step $\hor$. 

We may also consider identifying system parameters quickly, possibly in the presence of large, instantaneous disturbances to system parameters. We can accomplish this with an objective that minimizes stagewise uncertainty:
\begin{equation}
    \min_\pi \ 
    \mathbb{E}\left[\sum_{t=1}^\hor H(b_{\param, \hor})\right]
     \text{.}
\end{equation}

The goal in MIAC is to reduce uncertainty in the parameters and simultaneously control the system. Consider the system state $x$ and stagewise costs $C_t(x,a)$. The objective is: 
\begin{equation}
    \min_\pi \ 
    \mathbb{E}\left[\sum_{t=1}^\hor C_t(x_t, a_t) + \lambda H(b_{\param, \hor})\right]
     \text{,}
    \label{eq:miac_obj}
\end{equation}
with trade-off hyperparameter $\lambda$. That is, we wish to determine the policy $\pi$ that optimizes the control objective over the time horizon while reducing the uncertainty in the system parameters at the final time step.

\subsection{$\rho$-POMDP Formulation}
We propose solving the MIAC objective in \cref{eq:miac_obj} by formulating the problem as a $\rho$-POMDP that combines system states and parameters. Given a joint state $s = [x, \param]$, we can define belief-based rewards as the stagewise cost of the MIAC objective in \cref{eq:miac_obj}:
\begin{equation}
    R(b,a) = -\mathbb{E}_{x\sim b}[C(x,a)] - \lambda H(b_{\param}) \text{,}
\end{equation}
where $b_{\param}$ marginalizes the joint belief over $\param$. As is common in informative input design \cite{ott2023sequential}, we assume Gaussian beliefs. 

\begin{align}
    b(s) &= \mathcal{N} \left( s \mid \mu_s = 
    \begin{pmatrix}
        \mu_x \\ \mu_\param
    \end{pmatrix} , \Sigma_s = 
    \begin{pmatrix}
        \Sigma_{xx} & \Sigma_{x\param} \\ 
        \Sigma_{\param x} & \Sigma_{\param \param}
    \end{pmatrix} \right) \text{,}
\end{align}
with shorthand $b = [\mu_s \ \Sigma^\ast_s]^\top$,
where $\Sigma^\ast_s$ is the covariance matrix  $\Sigma_s$ flattened. Assuming nonlinear Gaussian dynamics and measurements, we write the joint transition function as:
\begin{align}
T(s' \mid s, a) &= \mathcal{N}\Biggl(
    \begin{pmatrix}
    x' \\ \param'
    \end{pmatrix}
    \mid \mu =
    \begin{pmatrix}
    f(x, a, \param) \\ \param
    \end{pmatrix}, \\ 
    &\quad \Sigma =
    \begin{pmatrix}
    W_x & 0 \\
    0 & W_\param
    \end{pmatrix}
    \Biggr)
\end{align}

where $f(x, a, \param)$ is the deterministic part of the dynamics function, and $W_x$ and $W_\param$ are process noise matrices for $x$ and $\param$ respectively. We can plan for disturbances to system parameters by assuming a nonzero $W_\param$. We will use $\bar{f}(s,a)$ as a shorthand for the deterministic part of the joint transition. 

We can also define a Gaussian observation function: 
\begin{align}
    O(o \mid s, a) &= O(o \mid x,  a, \param) \\
    &= \mathcal{N}(o \mid \mu = g(x,a), \Sigma = V) \text{,}
\end{align}
where $g(x, a)$ is the deterministic part of the observation function and $V$ is the observation noise. For fully observable systems, we approximate the true observation function \\$O(o \mid x, \param, a) = \delta(o=x)$ with a Gaussian centered at $x'$ with an arbitrarily small, user-defined covariance.  

\subsection{Belief-state iLQR}

We propose solving this reformulated MIAC problem by adapting Platt Jr. \textit{et al.}'s BiLQR approach to plan over system state and parameter beliefs with a receding horizon. 

% The first action from this plan is applied to the current state to obtain the next observation. An extended Kalman filter linearizes the dynamics and observation functions using the Jacobians for nonlinear systems and uses the observation to update the current belief. The loop repeats until the goal state is reached and the uncertainty is reduced below a certain tolerance, or the time horizon is reached. 

\begin{algorithm}
\caption{Belief-state iLQR planning applied to MIAC}
\label{alg:BiLQR}
\textbf{Require:} Nominal control sequence, $(\bar{a}_0, \dots, \bar{a}_{\hor-1})$, Current Belief State, $b_0 = b_0(x,\param) = [\mu_{x, 0}, \mu_{\param, 0}, \Sigma^*_{x, 0}]$

\begin{algorithmic}[1]
\State Set $\bar{b}_0 = b_0$ and $\delta a_t = 0$ for all $t \in \{0, \dots, \hor-1\}$
\While{not converged}
    \State \textit{Forward pass: } $\forall t \in \{0, \dots, \hor-1\}$
    \State Compute nominal trajectory $\bar{b}_{t+1} = F(\bar{b}_{t}, \bar{a}_t + \delta a_t)$
    \State Set $\bar{a}_t \gets \bar{a}_t + \delta a_t$ 
    \State Compute matrices $\tilde{A}_t$ and $\tilde{B}_t$ from \cref{eq:superAB}
    \State Compute reward, $R_t(b, a)$ around $\bar{b}_t, \bar{a}_t$
    \State \textit{Backward pass:} $\forall t \in \{\hor-1, \dots, 0\}$
    \State Update value approximation as in Eq. 15 of \cite{platt2010belief}
    \State Update feedback law $\delta a_t$ as in Eq. 16 of \cite{platt2010belief}
\EndWhile
\State \Return $\bar{a}_0$
\end{algorithmic}
\end{algorithm}
\cref{alg:BiLQR} depicts the adaptation of the BiLQR planner to MIAC. It initializes a nominal trajectory over a receding time horizon by forward-propagating the initial belief with zero control input. Then, it executes forward and backward passes, starting with the initial belief state to update the actions until the optimal sequence is found. In the initialization and forward pass, the belief-state dynamics \(F(b,a)\) propagate the belief state over the time horizon. The dynamics are
\begin{align}
b_{t+1} = & F(b_t,a_t) = 
\begin{pmatrix}
           \bar{f}(\mu_{s,t}, a_t)  \\
            \mu_{\param,t} \\
            \Sigma^*_{t+1} 
        \end{pmatrix} 
 \text{,} \\ 
 \Sigma_{t+1} = & (I - (A_t \Sigma_t A_t^\top + W_x) C_t^\top (C_t (A_t \Sigma_t A_t^\top \notag \\ 
 & + W_x) C_t^\top + V)^{-1} C_t ) (A_t \Sigma_t A_t^\top + W_x)\text{,}
    % & F(b_t,a_t) = \begin{pmatrix}
    %     \mu_{s,t+1} = f(\mu_{s,t}, a_t) \\ 
    %     \mu_{\param,t+1} = \mu_{\param,t} \\
    %     \Sigma_{t+1} = (I - (A_t \Sigma_t A_t^\top + W_x) C_t^\top (C_t (A_t \Sigma_t A_t^\top \\ + W_x) C_t^\top + V)^{-1} C_t ) (A_t \Sigma_t A_t^\top + W_x) 
    % \end{pmatrix}\text{,}
\end{align}
where $A_t$ is the joint dynamics Jacobian matrix and $C_t$ is the observation Jacobian matrix. The belief-state dynamics are linearized around the nominal trajectory. The linearized belief-state space matrices $\tilde{A}$ and $\tilde{B}$ are
\begin{equation}
    \begin{aligned}
        \tilde{A}_t &= \frac{\partial F}{\partial b} \mid_{(\Bar{b}_t, \Bar{a}_t)}  = \begin{pmatrix}
            \frac{\partial f}{\partial x_t} & 0 & 0 \\ 0 & I & 0 \\ 
        \frac{\partial \Sigma^\ast_t}{\partial x_t} & \frac{\partial \Sigma^\ast_t}{\partial \param_t} & \frac{\partial \Sigma^\ast_t}{\partial \Sigma_t}
        \end{pmatrix} \\
        \tilde{B}_t &= \frac{\partial F}{\partial a} \mid_{(\Bar{b}_t, \Bar{a}_t)} = \begin{pmatrix}
           \frac{\partial f}{\partial a_t}  \\
            0 \\
            0 
        \end{pmatrix} \text{,}
        \label{eq:superAB}
    \end{aligned}
\end{equation}
where $\Sigma^\ast_t$ is the stacked column representation of the covariance matrix $\Sigma$. The forward pass propagates the mean of the belief forward with the linearized state dynamics and the covariance through an extended Kalman filter assuming maximum-likelihood observations \(O = g(x,a)\).

The backward pass refines the policy using the quadratic reward that updates the optimal actions from the initial guess. In maximizing the reward, the resulting plan better aligns with the system dynamics and mitigates uncertainty. 

If only system identification is considered, the goal is to reduce the uncertainty in the system parameters. Thus, the per-time-step reward function is 
\begin{align}
R_t(b, a) = \mathbf{1}(t=\hor){\Sigma^\ast}_{\param\param,t}^\top \Lambda \Sigma^\ast_{\param\param,t} \text{,}
\label{eq:reward_bilqr}
\end{align}
where $\Lambda$ is a negative semi-definite matrix that penalizes uncertainty in system parameters at time $\hor$, measured by $\Sigma_{\param\param,\hor}$, the sub-vector of $\Sigma^\ast_t$ for just the covariance between the unknown system parameters. 

For MIAC, the stagewise reward function is 
\begin{align}
    R_t(b,a) = (\mu_{x,t} - \mu_{goal})^\top \hat{Q}_t (\mu_{x,t} - \mu_{goal}) \notag \\ + a_{t}^\top \hat{R}_t a_{t} + \mathbf{1}(t=\hor){\Sigma^\ast}_{\param\param,t}^\top \Lambda \Sigma^\ast_{\param\param,t} \text{,}
\end{align}
which incorporates negative semi-definite $\hat{Q}_t$ and $\hat{R}_t$ state and action cost matrices from traditional LQR. This formulation is equivalent to \cref{eq:miac_obj}, with $H(b_\param)$ as a quadratic cost function on system parameter uncertainty. The forward and backward passes are performed until the optimal control sequence converges. Only the first action from the plan is executed. We update the belief and replan after receiving new state information. Platt Jr. \textit{et al.} show that once the belief covariance is sufficiently reduced, the system enters a locally linear regime where LQR guarantees exponential convergence of the belief mean to the target. BiLQR, like iLQR, is not globally optimal but provides locally optimal solutions around a nominal belief trajectory.

The $\rho$-POMDP formulation of MIAC and the BiLQR planner allows efficient solutions in both fully and partially observable domains with continuous state spaces. 

\section{Experiments}
Our experiments consider both fully and partially observable systems to empirically demonstrate our adapted BiLQR's performance. As expected, performance degrades under partial observability compared to full observability, but BiLQR is more successful, and this case provides a strong lower-bound baseline for comparison. We compare informative input design performance against filtering and regression baselines and MIAC performance against model predictive control (MPC) with regression and extended Kalman filter (EKF) baselines. Moreover, we demonstrate robustness to time-varying disturbances in system parameters. 

We use the \texttt{POMDPs.jl} framework for our experiments~\cite{egorov2017pomdps}. For full experimental details, including cost matrices and hyperparameters, see \url{https://github.com/sisl/MIAC_BiLQR/}.

\subsection{Problem Domains}
We discuss the problem domains used in our experiments, the POMDP definition, their fully and partially observable variations, and the system parameters to be identified. 

\subsubsection{Cart-pole}
In this one-dimensional traditional control problem, a cart moves along a track, and a pole, attached by a pivot, must be balanced upright. The state of the system $x_t$ is described by the cart's position $p_t$, the pole's angle $\psi_t$, and their respective velocities $\dot{p}_t \text{ and } \dot{\psi}_t$. Control is achieved by applying a one-dimensional force $a$ to the cart \cite{barto1983neuronlike}. We assume the mass of the pole is unknown and seek to identify the log mass, $\param = \log m_p$. 

The dynamics follow $x_{t+1} = x_t + \dot{x}_t \delta t$ where
\begin{align}
    \dot{x}_t = 
 \begin{bmatrix}
    \dot{p}_t, \\
    \dot{\psi}_t, \\
    \frac{m_p \sin \psi_t \left( L \dot{\psi}_t^2 + g \cos \psi_t \right) + a}{h} \\
    -\frac{(m_c + m_p) g \sin \psi_t + m_p L \dot{\psi}_t^2 \sin \psi_t \cos \psi_t + a \cos \psi_t}{h L} 
\end{bmatrix}\text{.}
\end{align}
Reward is given for minimizing control effort and keeping the pole within 12$^{\circ}$ of the vertical \cite{barto1983neuronlike}. The partially observable setting includes noisy observations of cart and pole position, i.e., $g(x,a) = [p, \psi]^\top$ with Gaussian noise $V$.

\subsubsection{Aircraft Steady Flight}
We adapt the linear aircraft informative input design problem from Ott \textit{et al.} to a longitudinal motion model with three degrees of freedom. The goal is to identify the linear state-space dynamics defined by the transition matrix \(\Phi_1\) and input matrix \(\Phi_2\). The state \(x_t\) consists of horizontal and vertical velocity perturbations \(u_t \text{ and } w_t\), angle of attack \(\alpha_t\), and pitch rate \(\dot{\alpha}_t\). The action \(a_t\) includes elevator deflection \(\delta_e\) and throttle \(\delta_{th}\). To test the method on multiple unknown parameters while balancing computational efficiency, we treat the first columns of \(\Phi_1\) and \(\Phi_2\) as the unknown parameters.

The state evolution is governed by the equation:
\begin{equation*}
    x_{t+1} = \Phi_1 x_t + \Phi_2 a_t \text{.}
\end{equation*}
Reward is given for keeping $\alpha_t$ within $0^\circ$ to $30^\circ$, maintaining non-negative vertical velocity, and minimizing control effort. In the partially observable setting, all states except pitch rate are observed, i.e., $g(x,a) = [u, w, \alpha]^\top$ with Gaussian noise $V$.

\subsection{Experiments and Discussion}

\subsubsection{Informative Input Design Comparison}
\begin{table*}[htpb]
\ra{1.2}
\centering
\scalebox{0.9}{%
\begin{tabular}{@{}r r r r r r r r r @{}}
\toprule

& \multicolumn{2}{c}{\underline{\quad\quad\textbf{Cart-Pole Full Obs}\quad\quad}}
& \multicolumn{2}{c}{\underline{\quad\quad\textbf{Cart-Pole Partial Obs}\quad\quad}}
& \multicolumn{2}{c}{\underline{\quad\quad\quad\textbf{Aircraft Full Obs}\quad\quad\quad}}
& \multicolumn{2}{c}{\underline{\quad\quad\quad\textbf{Aircraft Partial Obs}\quad\quad\quad}} \\

\textbf{Solver} &
\multicolumn{1}{c}{$\mathrm{tr}\!\bigl(\Sigma_{\param \param,\hor}\bigr)$} &
\multicolumn{1}{c}{$\log p_{\hat{\param},\hor}$} &
\multicolumn{1}{c}{$\mathrm{tr}\!\bigl(\Sigma_{\param \param,\hor}\bigr)$} &
\multicolumn{1}{c}{$\log p_{\hat{\param},\hor}$} &
\multicolumn{1}{c}{$\mathrm{tr}\!\bigl(\Sigma_{\param \param,\hor}\bigr)$} &
\multicolumn{1}{c}{$\log p_{\hat{\param},\hor}$} &
\multicolumn{1}{c}{$\mathrm{tr}\!\bigl(\Sigma_{\param \param,\hor}\bigr)$} &
\multicolumn{1}{c}{$\log p_{\hat{\param},\hor}$} \\

\midrule

\texttt{BiLQR} &
$\mathbf{0.012 {\scriptstyle \pm 0.001}}$ &
$\mathbf{0.514 {\scriptstyle \pm 0.147}}$ &
$\mathbf{0.108 {\scriptstyle \pm 0.007}}$ &
$\mathbf{-2.396 {\scriptstyle \pm 1.824}}$ &
$\mathbf{7.994 {\scriptstyle \pm 0.470}}$ &
$\mathbf{-4.976 {\scriptstyle \pm 0.425}}$ &
$\mathbf{22.604 {\scriptstyle \pm 0.816}}$ &
$\mathbf{-7.792 {\scriptstyle \pm 1.202}}$ \\

\texttt{Random + EKF} &
$0.022 {\scriptstyle \pm 0.002}$ &
$-39.945 {\scriptstyle \pm 29.861}$ &
$0.132 {\scriptstyle \pm 0.011}$ &
$-4.507 {\scriptstyle \pm 2.932}$ &
$32.329 {\scriptstyle \pm 0.148}$ &
$-31.219 {\scriptstyle \pm 2.613}$ &
$45.970 {\scriptstyle \pm 0.079}$ &
$-17.837 {\scriptstyle \pm 1.937}$ \\

\texttt{Regression} &
$0.172 {\scriptstyle \pm 0.081}$ &
$-35.235 {\scriptstyle \pm 3.966}$ &
$-$ & $-$ &
$49.480 {\scriptstyle \pm 0.234}$ &
$-28.865 {\scriptstyle \pm 1.362}$ &
$-$ & $-$ \\

\bottomrule
\end{tabular}
}

\caption{Informative input design with $\rho$-POMDP planning, reported mean and standard error of $\Sigma_{\param\param, T}$ and $\log p_{\hat{\param},\hor}$ over the predicted distribution across 150 simulations using BiLQR with a covariance minimizing objective, random policy with EKF, and linear regression.}
\label{table:sysid_results}
% \vspace{-0.6cm}
\end{table*}

% A value of zero for $\Sigma_{\param\param, T}$ indicates having decreased our uncertainty fully in our system parameters. A higher $\log p_{\hat{\theta}, \hor}$ indicates a higher probability that our estimated value of the learned system parameter is accurate. 

We evaluate BiLQR on informative input design in fully and partially observable environments using (1) the trace of the final system parameter covariance, \(\Sigma_{\param\param, T}\) (zero indicates complete uncertainty reduction), and (2) the log likelihood of the true parameter, \(\log p_{\hat{\theta}, \tau}\) (higher implies more accurate estimates). We benchmark BiLQR against approximate least squares (regression) and an EKF applied to the joint state under a random policy sampled from \(\mathcal{U}(a_\text{min}, a_\text{max})\).

\Cref{table:sysid_results} summarizes the performance of the algorithms on the domains, averaged across 150 simulations per experiment. BiLQR significantly outperforms both baselines at reducing uncertainty in system parameters, as shown by the lower trace of the system parameter covariance matrix. Additionally, the log likelihood at the final time step for the system parameters is the highest for BiLQR, indicating a higher probability that the system parameters learned by BiLQR are accurate. In the partially observable environments, the uncertainty is higher, and the log likelihood is lower than in the fully observable counterparts, indicating that it was more difficult to identify the system parameters. Still, BiLQR outperformed the EKF baseline in this setting. 

% Note that linear regression could not be run on a partially observable environment. 

% \Cref{fig:bilqr_sysid_mp} shows the mean estimate and standard deviation of the pole mass over time using BiLQR, showing that the standard deviation is significantly reduced by the end of the 80-step time horizon. BiLQR was able to accurately identify the mass estimate by the 20th time step. 

% \begin{figure}[h!]
%     \centering
%     \resizebox{0.5\textwidth}{!}{\input{figs/bilqr_sysid_mp}}  % Path to the .tex file generated by PGFPlotsX
%     \caption{Mass estimate of the pole mass using BiLQR over 80 time steps for one example simulation.}
%      \label{fig:bilqr_sysid_mp}
% \end{figure}

\subsubsection{Model Identification Adaptive Control Comparison}

\begin{table*}[htpb]
\ra{1.2}
\centering
\scalebox{1.0}{
    % First table for Cart-Pole results
    \begin{tabular}{@{}c r r r r r r@{}}
      \toprule
      & \multicolumn{3}{c}{\textbf{Cart-Pole Full Obs}}
      & \multicolumn{3}{c}{\textbf{Cart-Pole Partial Obs}} \\
      \cmidrule(lr){2-4} \cmidrule(lr){5-7}
      % 
      % The first column is c (centered) for the "Solver" text and entries like BiLQR, 
      % the remaining six columns are 'r' for right-aligned numeric data:
      %
      \multicolumn{1}{c}{\textbf{Solver}} &
      \multicolumn{1}{c}{$\mathrm{tr}(\Sigma_{\param\param,\hor})$} &
      \multicolumn{1}{c}{$\log{p_{\hat{\param},\hor}}$} &
      \multicolumn{1}{c}{$R(b,a)$} &
      \multicolumn{1}{c}{$\mathrm{tr}(\Sigma_{\param\param,\hor})$} &
      \multicolumn{1}{c}{$\log{p_{\hat{\param},\hor}}$} &
      \multicolumn{1}{c}{$R(b,a)$} \\
      \midrule

        \texttt{BiLQR} &
        $\mathbf{0.010 \scriptstyle\pm 0.001}$ &
        $\mathbf{0.022 \scriptstyle\pm 0.328}$ & 
        $\mathbf{4.251 \scriptstyle\pm 2.050}$ &
        $\mathbf{0.140 \scriptstyle\pm 0.010}$ &
        $\mathbf{-0.699 \scriptstyle\pm 0.169}$ & 
        $\mathbf{4.035 \scriptstyle\pm 2.046}$ \\

        \texttt{MPC + Regression} &
        $0.096 \scriptstyle\pm 0.012 $ & 
        $-46.671 \scriptstyle\pm 6.270$ & 
        $3.687 \scriptstyle\pm 1.635 $ &
        - & - & - \\

        \texttt{MPC + EKF} &
        $0.011 \scriptstyle\pm 0.001 $ & 
        $-6.497 \scriptstyle\pm 5.102 $ & 
        $3.748 \scriptstyle\pm 2.185 $ &
        $0.145 \scriptstyle\pm 0.010$ & 
        $-13.114 \scriptstyle\pm 9.782$ & 
        $3.863 \scriptstyle\pm 2.052$ \\

        \texttt{Random + EKF} &
        $0.022 \scriptstyle\pm 0.002$ & 
        $-39.945 \scriptstyle\pm 29.861$ &  
        $3.584 \scriptstyle\pm 1.804$ &
        $0.132 \scriptstyle\pm 0.011 $ & 
        $-4.507 \scriptstyle\pm 2.932$ & 
        $3.606 \scriptstyle\pm 2.141$ \\

        \bottomrule
    \end{tabular}
}

\vspace{0.5cm}

\scalebox{1.0}{
    % Second table for Aircraft results
   \begin{tabular}{@{}c r r r r r r@{}}
      \toprule
      & \multicolumn{3}{c}{\textbf{Aircraft Full Obs}}
      & \multicolumn{3}{c}{\textbf{Aircraft Partial Obs}} \\
      \cmidrule(lr){2-4} \cmidrule(lr){5-7}
      % 
      % The first column is c (centered) for the "Solver" text and entries like BiLQR, 
      % the remaining six columns are 'r' for right-aligned numeric data:
      %
      \multicolumn{1}{c}{\textbf{Solver}} &
      \multicolumn{1}{c}{$\mathrm{tr}(\Sigma_{\param\param,\hor})$} &
      \multicolumn{1}{c}{$\log{p_{\hat{\param},\hor}}$} &
      \multicolumn{1}{c}{$R(b,a)$} &
      \multicolumn{1}{c}{$\mathrm{tr}(\Sigma_{\param\param,\hor})$} &
      \multicolumn{1}{c}{$\log{p_{\hat{\param},\hor}}$} &
      \multicolumn{1}{c}{$R(b,a)$} \\
      \midrule
    
        \texttt{BiLQR} &
        $\mathbf{2.057 \scriptstyle\pm 0.120}$ & 
        $\mathbf{-3.377 \scriptstyle\pm 3.353}$ & 
        $\mathbf{77.953 \scriptstyle \pm 12.905}$ &
        $\mathbf{22.609 \scriptstyle\pm 0.334}$ &
        $\mathbf{-7.919 \scriptstyle\pm 0.900}$ & 
        $\mathbf{78.429 \scriptstyle \pm 12.471}$ \\

        \texttt{MPC + Regression} &
        $44.974 \scriptstyle\pm 0.289$ & 
        $-180.014 \scriptstyle \pm 10.128$ &
        $3.707 \scriptstyle \pm 0.902 $ &
        $ - $ & $ - $ & $-$ \\

        \texttt{MPC + EKF} &
        $9.440 \scriptstyle\pm 0.212 $ & 
        $-7.578 \scriptstyle\pm 3.010$ & 
        $56.040 \scriptstyle \pm 18.239$ &
        $20.741 \scriptstyle\pm 0.146$ & 
        $-8.035 \scriptstyle\pm 6.288$ & 
        $58.033 \scriptstyle \pm 19.242$ \\

        \texttt{Random + EKF} &
        $32.329 \scriptstyle\pm 0.148$ & 
        $-31.219 \scriptstyle\pm 2.613$ & 
        $51.707 \scriptstyle \pm 17.305$ &
        $45.970 \scriptstyle\pm 0.079$ & 
        $-17.837 \scriptstyle\pm 1.937$ & 
        $51.967 \scriptstyle \pm 16.678$ \\

        \bottomrule
    \end{tabular}
}
\caption{MIAC with $\rho$-POMDP planning, reported mean and standard error of $\Sigma_{\param\param, T}$ and $\log p_{\hat{\param},\hor}$ over 150 simulations using BiLQR with a covariance-minimizing objective, MPC with linear regression and EKF, and a random policy with EKF.}

\label{table:miac_results}
\vspace{-1em}
% \vspace{-0.6cm}
\end{table*}

% \begin{table*}[htpb]
% \ra{1.2}
% \centering
% \scalebox{1.0}{
% \input{tables/miac_results}
% }
% \caption{MIAC with $\rho$-POMDP planning demonstrations, reported metric mean and standard deviation across 150 simulations using BiLQR with a covariance minimizing objective, MPC with linear regression and EKF, and a random policy with EKF.}
% \label{table:miac_results}
% % \vspace{-0.6cm}
% \end{table*}

Our evaluation metrics for MIAC are the trace of $\Sigma_{\param\param, T}$, $\log p_{\hat{\theta}, \hor}$, and the expected reward over the time horizon $\mathbb{E}[R(b,a)]$, indicating success in achieving the control objective. For the cart-pole domain, the objective is to balance the pole at $\psi = \pi/2$. For the aircraft domain, the objective is to reach a forward speed of 100 m/s and reduce vertical velocity, angle of attack, and pitch angle rate to zero. We benchmark BiLQR against both a least squares approximation and using an EKF when following model predictive control (by solving a quadratic program) and also using an EKF when following a random policy.

\Cref{table:miac_results} summarizes the performance of the algorithms on the different domains, averaged across 150 simulations per domain and observability variation. In both problem domains with both full and partial observability, we see that BiLQR reduces uncertainty, as shown by the low trace of the system parameter covariance matrix. It performed comparably to MPC with EKF updates, indicating it learned the system parameters just as effectively. Also, as indicated by the higher average reward per time step, BiLQR drives the system to achieve the objective better than all of the baselines. 

\begin{figure}[h!]
    \vspace{-1em}
    \centering
    \resizebox{0.5\textwidth}{!}{% Recommended preamble:
% \usetikzlibrary{arrows.meta}
% \usetikzlibrary{backgrounds}
% \usepgfplotslibrary{patchplots}
% \usepgfplotslibrary{fillbetween}
% \pgfplotsset{%
%     layers/standard/.define layer set={%
%         background,axis background,axis grid,axis ticks,axis lines,axis tick labels,pre main,main,axis descriptions,axis foreground%
%     }{
%         grid style={/pgfplots/on layer=axis grid},%
%         tick style={/pgfplots/on layer=axis ticks},%
%         axis line style={/pgfplots/on layer=axis lines},%
%         label style={/pgfplots/on layer=axis descriptions},%
%         legend style={/pgfplots/on layer=axis descriptions},%
%         title style={/pgfplots/on layer=axis descriptions},%
%         colorbar style={/pgfplots/on layer=axis descriptions},%
%         ticklabel style={/pgfplots/on layer=axis tick labels},%
%         axis background@ style={/pgfplots/on layer=axis background},%
%         3d box foreground style={/pgfplots/on layer=axis foreground},%
%     },
% }

\begin{tikzpicture}[/tikz/background rectangle/.style={fill={rgb,1:red,1.0;green,1.0;blue,1.0}, fill opacity={1.0}, draw opacity={1.0}}, show background rectangle]
\begin{axis}[point meta max={nan}, point meta min={nan}, legend cell align={left}, legend columns={1}, title={}, title style={at={{(0.5,1)}}, anchor={south}, font={{\fontsize{14 pt}{18.2 pt}\selectfont}}, color={rgb,1:red,0.0;green,0.0;blue,0.0}, draw opacity={1.0}, rotate={0.0}, align={center}}, legend style={color={rgb,1:red,0.0;green,0.0;blue,0.0}, draw opacity={1.0}, line width={1}, solid, fill={rgb,1:red,1.0;green,1.0;blue,1.0}, fill opacity={1.0}, text opacity={1.0}, font={{\fontsize{14 pt}{10.4 pt}\selectfont}}, text={rgb,1:red,0.0;green,0.0;blue,0.0}, cells={anchor={center}}, at={(0.02, 0.02)}, anchor={south west}}, axis background/.style={fill={rgb,1:red,1.0;green,1.0;blue,1.0}, opacity={1.0}}, anchor={north west}, xshift={1.0mm}, yshift={-1.0mm}, width={150.4mm}, height={99.6mm}, scaled x ticks={false}, xlabel={Time Step}, x tick style={color={rgb,1:red,0.0;green,0.0;blue,0.0}, opacity={1.0}}, x tick label style={color={rgb,1:red,0.0;green,0.0;blue,0.0}, opacity={1.0}, rotate={0}}, xlabel style={at={(ticklabel cs:0.5)}, anchor=near ticklabel, at={{(ticklabel cs:0.5)}}, anchor={near ticklabel}, font={{\fontsize{14 pt}{14.3 pt}\selectfont}}, color={rgb,1:red,0.0;green,0.0;blue,0.0}, draw opacity={1.0}, rotate={0.0}}, xmajorgrids={true}, xmin={0.129999999999999}, xmax={30.87}, xticklabels={{$5$,$10$,$15$,$20$,$25$,$30$}}, xtick={{5.0,10.0,15.0,20.0,25.0,30.0}}, xtick align={inside}, xticklabel style={font={{\fontsize{14 pt}{10.4 pt}\selectfont}}, color={rgb,1:red,0.0;green,0.0;blue,0.0}, draw opacity={1.0}, rotate={0.0}}, x grid style={color={rgb,1:red,0.0;green,0.0;blue,0.0}, draw opacity={0.1}, line width={0.5}, solid}, axis x line*={left}, x axis line style={color={rgb,1:red,0.0;green,0.0;blue,0.0}, draw opacity={1.0}, line width={1}, solid}, scaled y ticks={false}, ylabel={$\log p(\hat{\param} \mid a_{1:t}, o_{1:t})$}, y tick style={color={rgb,1:red,0.0;green,0.0;blue,0.0}, opacity={1.0}}, y tick label style={color={rgb,1:red,0.0;green,0.0;blue,0.0}, opacity={1.0}, rotate={0}}, ylabel style={at={(ticklabel cs:0.5)}, anchor=near ticklabel, at={{(ticklabel cs:0.5)}}, anchor={near ticklabel}, font={{\fontsize{14 pt}{14.3 pt}\selectfont}}, color={rgb,1:red,0.0;green,0.0;blue,0.0}, draw opacity={1.0}, rotate={0.0}}, ymajorgrids={true}, ymin={-16.478542776556182}, ymax={1.4878572838691824}, yticklabels={{$-15$,$-10$,$-5$,$0$}}, ytick={{-15.0,-10.0,-5.0,0.0}}, ytick align={inside}, yticklabel style={font={{\fontsize{14 pt}{10.4 pt}\selectfont}}, color={rgb,1:red,0.0;green,0.0;blue,0.0}, draw opacity={1.0}, rotate={0.0}}, y grid style={color={rgb,1:red,0.0;green,0.0;blue,0.0}, draw opacity={0.1}, line width={0.5}, solid}, axis y line*={left}, y axis line style={color={rgb,1:red,0.0;green,0.0;blue,0.0}, draw opacity={1.0}, line width={1}, solid}, colorbar={false}]
    \addplot[color={rgb,1:red,0.0;green,0.6056;blue,0.9787}, name path={1}, draw opacity={1.0}, line width={1}, solid]
        table[row sep={\\}]
        {
            \\
            2.0  -1.2655122960795169  \\
            3.0  -1.2495449880036364  \\
            4.0  -1.2599730026982172  \\
            5.0  -1.025551537425474  \\
            6.0  -0.3886399842121346  \\
            7.0  0.35795993167500006  \\
            8.0  0.36857899621959533  \\
            9.0  0.36876748246469054  \\
            10.0  0.3694835467308996  \\
            11.0  0.3738131538070652  \\
            12.0  0.37633761516292946  \\
            13.0  0.37609107878170805  \\
            14.0  0.38160410247636195  \\
            15.0  0.38819349159976246  \\
            16.0  0.07087389309284198  \\
            17.0  -0.1619674284700915  \\
            18.0  0.6701127550930891  \\
            19.0  0.8326450386044707  \\
            20.0  0.8400546651730002  \\
            21.0  0.8929131606209518  \\
            22.0  0.8681151153314706  \\
            23.0  0.8871957555861248  \\
            24.0  0.9102020489764207  \\
            25.0  0.8282123138474843  \\
            26.0  0.8816631380032264  \\
            27.0  0.8652382955174833  \\
            28.0  0.9112896461484166  \\
            29.0  0.9386870059311294  \\
            30.0  0.9793742632911064  \\
        }
        ;
    \addlegendentry {BiLQR}
    \addplot[color={rgb,1:red,0.8889;green,0.4356;blue,0.2781}, name path={2}, draw opacity={1.0}, line width={1}, solid]
        table[row sep={\\}]
        {
            \\
            2.0  -1.2654758512594688  \\
            3.0  -1.2664431553957498  \\
            4.0  -1.1451873281478688  \\
            5.0  -0.5882532651722325  \\
            6.0  -0.5973492810726586  \\
            7.0  -0.3358252279860354  \\
            8.0  -0.412968272802959  \\
            9.0  -0.024286146040801204  \\
            10.0  -0.09723600685366374  \\
            11.0  -0.11871155998426325  \\
            12.0  -0.1258625353848587  \\
            13.0  -0.1244906888388369  \\
            14.0  -0.07358622740012971  \\
            15.0  -0.014789727291566745  \\
            16.0  -0.10435822375179382  \\
            17.0  0.01327237357517419  \\
            18.0  0.01570954685045972  \\
            19.0  0.03157881106556437  \\
            20.0  0.045821104606899477  \\
            21.0  0.05668823072718726  \\
            22.0  0.021865460012826354  \\
            23.0  0.2319919043106386  \\
            24.0  0.2365792603712995  \\
            25.0  -0.03487328929893474  \\
            26.0  -0.32078482042887224  \\
            27.0  -0.26777762021264917  \\
            28.0  -0.2703841774661714  \\
            29.0  -0.27647594809586484  \\
            30.0  -0.2938271597087432  \\
        }
        ;
    \addlegendentry {Random + EKF}
    \addplot[color={rgb,1:red,0.2422;green,0.6433;blue,0.3044}, name path={3}, draw opacity={1.0}, line width={1}, solid]
        table[row sep={\\}]
        {
            \\
            2.0  -1.2655120947784013  \\
            3.0  -1.2648585485855428  \\
            4.0  -1.2566782548125464  \\
            5.0  -0.6951913638246447  \\
            6.0  -0.8454205393846366  \\
            7.0  -0.49549155545271806  \\
            8.0  -1.1307084425340963  \\
            9.0  -0.9429922424325728  \\
            10.0  -0.843357425204674  \\
            11.0  -0.9512462224344316  \\
            12.0  -1.105330304277645  \\
            13.0  -1.0328497238904644  \\
            14.0  -1.003553797001756  \\
            15.0  -0.9677988776233304  \\
            16.0  -0.849851028528106  \\
            17.0  -0.5693214030763121  \\
            18.0  -0.9302484134919178  \\
            19.0  -2.3596152411208395  \\
            20.0  -1.3169662294645303  \\
            21.0  -0.9161321377470324  \\
            22.0  -0.9750771337898436  \\
            23.0  -1.0579341797010817  \\
            24.0  -1.0983077718034417  \\
            25.0  -1.1808790771532154  \\
            26.0  -1.179424199721378  \\
            27.0  -1.1652615923310672  \\
            28.0  -1.177863669214408  \\
            29.0  -1.1151237608515434  \\
            30.0  -1.2770541072311443  \\
        }
        ;
    \addlegendentry {MPC + EKF}
    \addplot[color={rgb,1:red,0.7644;green,0.4441;blue,0.8243}, name path={4}, draw opacity={1.0}, line width={1}, solid]
        table[row sep={\\}]
        {
            \\
            2.0  -1.2655120947784013  \\
            3.0  -0.6914439591736582  \\
            4.0  -2.5519247588907685  \\
            5.0  -3.9184908701350665  \\
            6.0  -4.339953040284092  \\
            7.0  -4.220445568646506  \\
            8.0  -4.875992717819889  \\
            9.0  -5.007774529971049  \\
            10.0  -5.06282195903667  \\
            11.0  -5.511323962139108  \\
            12.0  -5.756008146522032  \\
            13.0  -5.090949875023587  \\
            14.0  -2.556174207597563  \\
            15.0  -6.815440806683252  \\
            16.0  -5.500001213131399  \\
            17.0  -5.692164177798747  \\
            18.0  -5.966782378522645  \\
            19.0  -6.033625085288099  \\
            20.0  -6.322633339772843  \\
            21.0  -6.028447733302162  \\
            22.0  -6.166605809201569  \\
            23.0  -4.520400793730662  \\
            24.0  -4.914944916341645  \\
            25.0  -4.730928667185697  \\
            26.0  -5.449701037539428  \\
            27.0  -7.047296576239198  \\
            28.0  -15.360550004221832  \\
            29.0  -15.802785368032602  \\
            30.0  -15.970059755978106  \\
        }
        ;
    \addlegendentry {MPC + Regression}
\end{axis}
\end{tikzpicture}}  % Path to the .tex file generated by PGFPlotsX
    \caption{MIAC $\log p(\hat{\param} \mid a_{1:t}, o_{1:t})$ for BiLQR, Random + EKF, MPC + EKF, and MPC + Regression in one example simulation of the fully observable cart-pole environment (showing the mean trend with error would obscure the BiLQR trend, since the baseline uncertainties in $\log p_{\hat{\param},\tau}$ are orders of magnitude larger than BiLQR).}
    \vspace{-1em}
     \label{fig:fullobs_cartpole}
\end{figure}
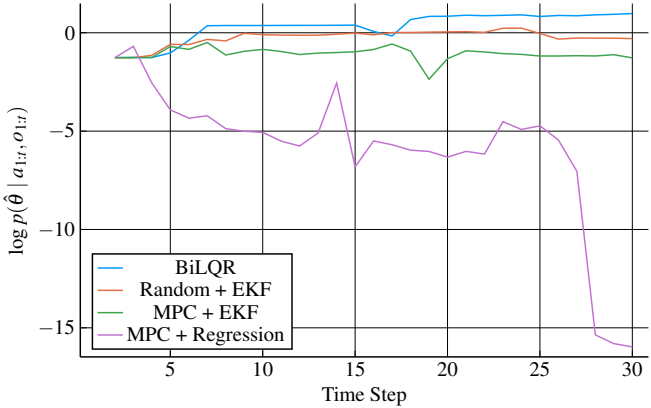

\Cref{fig:fullobs_cartpole} shows performance over time for the cart-pole fully observable environment. BiLQR does better than EKF and both MPC baselines in reducing the uncertainty about the pole mass, as shown by the higher final log likelihood. This measure indicates that the BiLQR estimate achieved for the mass at the final time step is most accurate. In this plot, BiLQR’s log likelihood remains higher than the other baselines for most of the time horizon, indicating that it finds more accurate estimates earlier than the baselines.

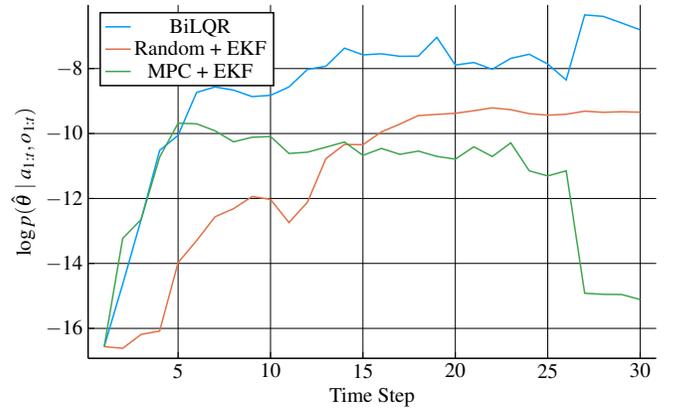
\begin{figure}[h!]
    \centering
    \resizebox{0.5\textwidth}{!}{% Recommended preamble:
% \usetikzlibrary{arrows.meta}
% \usetikzlibrary{backgrounds}
% \usepgfplotslibrary{patchplots}
% \usepgfplotslibrary{fillbetween}
% \pgfplotsset{%
%     layers/standard/.define layer set={%
%         background,axis background,axis grid,axis ticks,axis lines,axis tick labels,pre main,main,axis descriptions,axis foreground%
%     }{
%         grid style={/pgfplots/on layer=axis grid},%
%         tick style={/pgfplots/on layer=axis ticks},%
%         axis line style={/pgfplots/on layer=axis lines},%
%         label style={/pgfplots/on layer=axis descriptions},%
%         legend style={/pgfplots/on layer=axis descriptions},%
%         title style={/pgfplots/on layer=axis descriptions},%
%         colorbar style={/pgfplots/on layer=axis descriptions},%
%         ticklabel style={/pgfplots/on layer=axis tick labels},%
%         axis background@ style={/pgfplots/on layer=axis background},%
%         3d box foreground style={/pgfplots/on layer=axis foreground},%
%     },
% }

\begin{tikzpicture}[/tikz/background rectangle/.style={fill={rgb,1:red,1.0;green,1.0;blue,1.0}, fill opacity={1.0}, draw opacity={1.0}}, show background rectangle]
\begin{axis}[point meta max={nan}, point meta min={nan}, legend cell align={left}, legend columns={1}, title={}, title style={at={{(0.5,1)}}, anchor={south}, font={{\fontsize{14 pt}{18.2 pt}\selectfont}}, color={rgb,1:red,0.0;green,0.0;blue,0.0}, draw opacity={1.0}, rotate={0.0}, align={center}},  legend style={
        at={(0.02, 0.98)}, % move legend to top left corner
        anchor=north west,
        color={rgb,1:red,0.0;green,0.0;blue,0.0},
        draw opacity={1.0},
        line width={1},
        solid,
        fill={rgb,1:red,1.0;green,1.0;blue,1.0},
        fill opacity={1.0},
        text opacity={1.0},
        font={{\fontsize{14 pt}{10.4 pt}\selectfont}},
        text={rgb,1:red,0.0;green,0.0;blue,0.0},
        cells={anchor=center},
    }, axis background/.style={fill={rgb,1:red,1.0;green,1.0;blue,1.0}, opacity={1.0}}, anchor={north west}, xshift={1.0mm}, yshift={-1.0mm}, width={150.4mm}, height={99.6mm}, scaled x ticks={false}, xlabel={Time Step}, x tick style={color={rgb,1:red,0.0;green,0.0;blue,0.0}, opacity={1.0}}, x tick label style={color={rgb,1:red,0.0;green,0.0;blue,0.0}, opacity={1.0}, rotate={0}}, xlabel style={at={(ticklabel cs:0.5)}, anchor=near ticklabel, at={{(ticklabel cs:0.5)}}, anchor={near ticklabel}, font={{\fontsize{14 pt}{14.3 pt}\selectfont}}, color={rgb,1:red,0.0;green,0.0;blue,0.0}, draw opacity={1.0}, rotate={0.0}}, xmajorgrids={true}, xmin={0.129999999999999}, xmax={30.87}, xticklabels={{$5$,$10$,$15$,$20$,$25$,$30$}}, xtick={{5.0,10.0,15.0,20.0,25.0,30.0}}, xtick align={inside}, xticklabel style={font={{\fontsize{14 pt}{10.4 pt}\selectfont}}, color={rgb,1:red,0.0;green,0.0;blue,0.0}, draw opacity={1.0}, rotate={0.0}}, x grid style={color={rgb,1:red,0.0;green,0.0;blue,0.0}, draw opacity={0.1}, line width={0.5}, solid}, axis x line*={left}, x axis line style={color={rgb,1:red,0.0;green,0.0;blue,0.0}, draw opacity={1.0}, line width={1}, solid}, scaled y ticks={false}, ylabel={$\log p(\hat{\param} \mid a_{1:t}, o_{1:t})$}, y tick style={color={rgb,1:red,0.0;green,0.0;blue,0.0}, opacity={1.0}}, y tick label style={color={rgb,1:red,0.0;green,0.0;blue,0.0}, opacity={1.0}, rotate={0}}, ylabel style={at={(ticklabel cs:0.5)}, anchor=near ticklabel, at={{(ticklabel cs:0.5)}}, anchor={near ticklabel}, font={{\fontsize{14 pt}{14.3 pt}\selectfont}}, color={rgb,1:red,0.0;green,0.0;blue,0.0}, draw opacity={1.0}, rotate={0.0}}, ymajorgrids={true}, ymin={-16.92160078101308}, ymax={-6.040347774416066}, yticklabels={{$-16$,$-14$,$-12$,$-10$,$-8$}}, ytick={{-16.0,-14.0,-12.0,-10.0,-8.0}}, ytick align={inside}, yticklabel style={font={{\fontsize{14 pt}{10.4 pt}\selectfont}}, color={rgb,1:red,0.0;green,0.0;blue,0.0}, draw opacity={1.0}, rotate={0.0}}, y grid style={color={rgb,1:red,0.0;green,0.0;blue,0.0}, draw opacity={0.1}, line width={0.5}, solid}, axis y line*={left}, y axis line style={color={rgb,1:red,0.0;green,0.0;blue,0.0}, draw opacity={1.0}, line width={1}, solid}, colorbar={false}]
    \addplot[color={rgb,1:red,0.0;green,0.6056;blue,0.9787}, name path={5}, draw opacity={1.0}, line width={1}, solid]
        table[row sep={\\}]
        {
            \\
            1.0  -16.56184863761357  \\
            2.0  -14.64133480987622  \\
            3.0  -12.644899160333997  \\
            4.0  -10.520920670617793  \\
            5.0  -10.053290157279548  \\
            6.0  -8.735008682182555  \\
            7.0  -8.565732192272044  \\
            8.0  -8.663962329091325  \\
            9.0  -8.862746021121808  \\
            10.0  -8.824429479818079  \\
            11.0  -8.566478072870273  \\
            12.0  -8.024599660047365  \\
            13.0  -7.926845864548692  \\
            14.0  -7.371161164305208  \\
            15.0  -7.57831645325153  \\
            16.0  -7.545187053228206  \\
            17.0  -7.622281035024189  \\
            18.0  -7.618322950136208  \\
            19.0  -7.037985192141512  \\
            20.0  -7.89129865533518  \\
            21.0  -7.8140093408833815  \\
            22.0  -8.020483918932086  \\
            23.0  -7.686400159504405  \\
            24.0  -7.559888373421296  \\
            25.0  -7.860099342413554  \\
            26.0  -8.351371752956487  \\
            27.0  -6.348307765168813  \\
            28.0  -6.393275822286334  \\
            29.0  -6.596395253841947  \\
            30.0  -6.8066259455669655  \\
        }
        ;
    \addlegendentry {BiLQR}
    \addplot[color={rgb,1:red,0.8889;green,0.4356;blue,0.2781}, name path={6}, draw opacity={1.0}, line width={1}, solid]
        table[row sep={\\}]
        {
            \\
            1.0  -16.56184863761357  \\
            2.0  -16.613640790260334  \\
            3.0  -16.185179855851253  \\
            4.0  -16.0821892024426  \\
            5.0  -13.975608497470903  \\
            6.0  -13.297021611074593  \\
            7.0  -12.563437876729413  \\
            8.0  -12.313750145499805  \\
            9.0  -11.940460856869933  \\
            10.0  -12.029139751710813  \\
            11.0  -12.741767035617153  \\
            12.0  -12.11330037907949  \\
            13.0  -10.771154702830668  \\
            14.0  -10.33664186462149  \\
            15.0  -10.344007136479751  \\
            16.0  -9.949481267204394  \\
            17.0  -9.708582723477754  \\
            18.0  -9.447189761486603  \\
            19.0  -9.41079087566195  \\
            20.0  -9.376351767957546  \\
            21.0  -9.298346502162332  \\
            22.0  -9.209054125552655  \\
            23.0  -9.266543596907951  \\
            24.0  -9.389079469700722  \\
            25.0  -9.435902288687382  \\
            26.0  -9.407045031749435  \\
            27.0  -9.311285708143693  \\
            28.0  -9.34703444389769  \\
            29.0  -9.329477834161034  \\
            30.0  -9.34370234222312  \\
        }
        ;
    \addlegendentry {Random + EKF}
    \addplot[color={rgb,1:red,0.2422;green,0.6433;blue,0.3044}, name path={7}, draw opacity={1.0}, line width={1}, solid]
        table[row sep={\\}]
        {
            \\
            1.0  -16.56184863761357  \\
            2.0  -13.231303153348453  \\
            3.0  -12.64718818415266  \\
            4.0  -10.730386153196706  \\
            5.0  -9.682377271029168  \\
            6.0  -9.700944818886514  \\
            7.0  -9.917175778003077  \\
            8.0  -10.2556202462249  \\
            9.0  -10.115834959234594  \\
            10.0  -10.092573410380666  \\
            11.0  -10.614591388569067  \\
            12.0  -10.574496232466817  \\
            13.0  -10.420520369476087  \\
            14.0  -10.261049040155898  \\
            15.0  -10.672824269497156  \\
            16.0  -10.459199752465796  \\
            17.0  -10.640691618061371  \\
            18.0  -10.540987352718375  \\
            19.0  -10.704907304893553  \\
            20.0  -10.788163366568787  \\
            21.0  -10.408648735679684  \\
            22.0  -10.7081561461832  \\
            23.0  -10.286487462929596  \\
            24.0  -11.14858101301449  \\
            25.0  -11.30607037908842  \\
            26.0  -11.147879393286118  \\
            27.0  -14.918412488748473  \\
            28.0  -14.951345686202082  \\
            29.0  -14.95705712409511  \\
            30.0  -15.111535861848488  \\
        }
        ;
    \addlegendentry {MPC + EKF}
\end{axis}
\end{tikzpicture}}  % Path to the .tex file generated by PGFPlotsX
    \caption{MIAC $\log p(\hat{\param} \mid a_{1:t}, o_{1:t})$ for BiLQR, Random + EKF, and MPC + EKF in one example simulation of the partially observable aircraft environment. }
    \vspace{-1em}
    \label{fig:fullobs_xplane}
\end{figure}

\Cref{fig:fullobs_xplane} shows performance over time for the partially observable aircraft environment. BiLQR performs better than EKF and MPC with EKF in reducing the uncertainty about unknown elements of the linear state space \(\Phi_1\) and \(\Phi_2\) matrices, as shown by the higher log likelihood across most of the time horizon and at the final time step. This log likelihood was calculated from the multivariate Gaussian distribution defined by the vector of means for the unknown elements and the associated covariance matrix. BiLQR's likelihood also rises faster than the baselines, so more accurate estimates were found earlier in this example as well. 

% \begin{figure}[h!]
%     \centering
%     \resizebox{0.5\textwidth}{!}{\input{figs/accumulated_rewards_comparison}}  % Path to the .tex file generated by PGFPlotsX
%     \caption{MIAC Accumulated Reward over Time Horizon for BiLQR, Random + EKF, MPC + EKF, and MPC + Regression in one example simulation of the fully observable cart-pole environment.}
%     \label{fig:acc_reward}
% \end{figure}

% \begin{figure}[h!]
%     \centering
%     \resizebox{0.5\textwidth}{!}{\input{figs/reward_miac_plane}}  % Path to the .tex file generated by PGFPlotsX
%     \caption{MIAC Accumulated Reward over Time Horizon for BiLQR, Random + EKF, MPC + EKF, and MPC + Regression in one example simulation of the fully observable cart-pole environment.}
%     \label{fig:acc_reward}
% \end{figure}

% To analyze BiLQR's ability to control the system during MIAC, the accumulated reward over time for BiLQR, Random + EKF, MPC + EKF, and MPC + Regression was plotted (\cref{fig:acc_reward}). The reward was gained from balancing the pole within 12 degrees from the vertical \cite{barto1983neuronlike} and using minimal control effort. BiLQR achieves a higher accumulated award over most of the time horizon, outperforming the other baselines. 

\subsubsection{System Identification during Disturbances}

\begin{figure}[ht!]
    \centering
    \resizebox{0.5\textwidth}{!}{\input{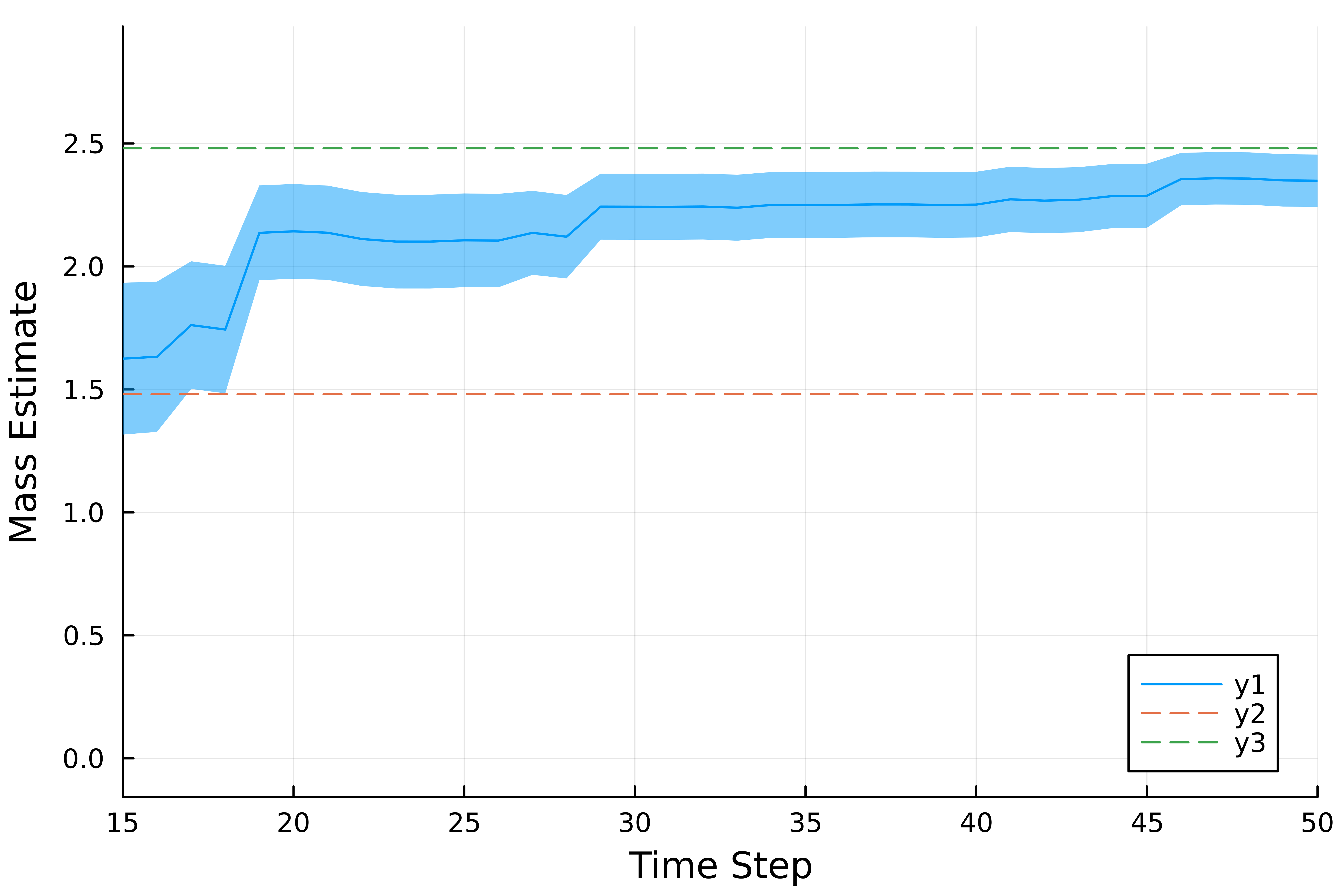}}  % Path to the .tex file generated by PGFPlotsX
    \caption{ Tracking change in pole mass that changes suddenly at time $t = t_c$ for one example simulation with BiLQR.}
    \vspace{-1em}
    \label{fig:timevar}
\end{figure}

In the $\rho$-POMDP formulation, a nonzero \(W_\param\) enables robust planning against parameter disturbances. For example, although the cart-pole mass remains constant in practice, a large disturbance may alter its modeled value to better model the dynamics. We simulate a sudden change in mass to test our approach with non-stationary parameters. As shown in \Cref{fig:timevar}, BiLQR quickly tracks the change and reduces uncertainty about the pole’s mass, with a slight delay to first detect and then adjust to the abrupt shift using observed data.
\section{Conclusion}

We framed informative input design and model identification adaptive control in fully and partially observable systems as a $\rho$-POMDP, with system parameters as hidden variables and a belief-based objective to balance control with system identification. Our experimental results on both fully observable and partially observable domains demonstrate that our adapted BiLQR significantly outperforms baselines, particularly in scenarios with high measurement noise. We also demonstrate robustness to time-varying system parameters.

Limitations of our approach include its local optimality and the computational challenges of scaling to high-dimensional state or parameter spaces. Real-time feasibility may be improved through warm-started linearizations or low-rank approximations. Framing MIAC as a $\rho$-POMDP also invites opportunities for incorporating constraints and exploring other belief-space planning approaches.

\bibliographystyle{IEEEtran}
\bibliography{sislstrings,references}

\begin{thebibliography}{10}
\providecommand{\url}[1]{#1}
\csname url@rmstyle\endcsname
\providecommand{\newblock}{\relax}
\providecommand{\bibinfo}[2]{#2}
\providecommand\BIBentrySTDinterwordspacing{\spaceskip=0pt\relax}
\providecommand\BIBentryALTinterwordstretchfactor{4}
\providecommand\BIBentryALTinterwordspacing{\spaceskip=\fontdimen2\font plus
\BIBentryALTinterwordstretchfactor\fontdimen3\font minus
  \fontdimen4\font\relax}
\providecommand\BIBforeignlanguage[2]{{%
\expandafter\ifx\csname l@#1\endcsname\relax
\typeout{** WARNING: IEEEtran.bst: No hyphenation pattern has been}%
\typeout{** loaded for the language `#1'. Using the pattern for}%
\typeout{** the default language instead.}%
\else
\language=\csname l@#1\endcsname
\fi
#2}}

\bibitem{ljung1999system}
L.~Ljung, \emph{System {Identification}: Theory for the User}.\hskip 1em plus
  0.5em minus 0.4em\relax Prentice Hall, 1999.

\bibitem{ott2024informative}
J.~Ott, M.~J. Kochenderfer, and S.~Boyd, ``Informative input design for dynamic
  mode decomposition,'' in \emph{Conference on Learning for Dynamics and
  Control (L4DC)}, 2025.

\bibitem{johansson2000statespace}
R.~Johansson, A.~Robertsson, K.~Nilsson, and M.~Verhaegen, ``State-space system
  identification of robot manipulator dynamics,'' \emph{Mechatronics}, vol.~10,
  no.~3, pp. 403--418, 2000.

\bibitem{wiens2015novel}
A.~D. Wiens and O.~T. Inan, ``A {novel} {system} {identification} {technique}
  for {improved} {wearable} {hemodynamics} {assessment},'' \emph{IEEE
  Transactions on Biomedical Engineering}, vol.~62, no.~5, pp. 1345--1354,
  2015.

\bibitem{los2006system}
C.~A. Los, ``System identification in noisy data environments: {An} application
  to six {Asian} stock markets,'' \emph{Journal of Banking \& Finance},
  vol.~30, no.~7, pp. 1997--2024, 2006.

\bibitem{oreg2019model}
Z.~Öreg, H.-S. Shin, and A.~Tsourdos, ``Model identification adaptive control:
  implementation case studies for a high manoeuvrability aircraft,'' in
  \emph{{Mediterranean} {Conference} on {Control} and {Automation}}, 2019, pp.
  559--564.

\bibitem{nishimura2021rat}
H.~Nishimura, N.~Mehr, A.~Gaidon, and M.~Schwager, ``{RAT} {iLQR}: {A} risk
  auto-tuning controller to optimally account for stochastic model mismatch,''
  \emph{IEEE Robotics and Automation Letters}, vol.~6, no.~2, pp. 763--770,
  2021.

\bibitem{wahlberg_optimal_2010}
\BIBentryALTinterwordspacing
B.~Wahlberg, H.~Hjalmarsson, and M.~Annergren, ``On optimal input design in
  system identification for control,'' in \emph{IEEE Conference on Decision and
  Control (CDC)}, 2010, pp. 5548--5553. [Online]. Available:
  \url{https://ieeexplore.ieee.org/abstract/document/5717863}
\BIBentrySTDinterwordspacing

\bibitem{slade2017simultaneous}
P.~Slade, P.~Culbertson, Z.~Sunberg, and M.~Kochenderfer, ``Simultaneous active
  parameter estimation and control using sampling-based {Bayesian}
  reinforcement learning,'' in \emph{IEEE/RSJ International Conference on
  Intelligent Robots and Systems (IROS)}, 2017, pp. 804--810.

\bibitem{araya2010pomdp}
M.~Araya, O.~Buffet, V.~Thomas, and F.~Charpillet, ``A {POMDP} extension with
  belief-dependent rewards,'' in \emph{Advances in Neural Information
  Processing Systems (NIPS)}, 2010.

\bibitem{platt2010belief}
R.~Platt~Jr., R.~Tedrake, L.~Kaelbling, and T.~Lozano-Perez, ``Belief space
  planning assuming maximum likelihood observations,'' in \emph{Robotics:
  Science and Systems}, 2010.

\bibitem{kopp1963linear}
R.~E. Kopp and R.~J. Orford, ``{Linear} {regression} {applied} {to} {system}
  {identification} {for} {adaptive} {control} {systems},'' \emph{AIAA Journal},
  vol.~1, no.~10, pp. 2300--2306, 1963.

\bibitem{gedon2021deep}
D.~Gedon, N.~Wahlström, T.~B. Schön, and L.~Ljung, ``Deep state space models
  for nonlinear system identification,'' in \emph{International Federation of
  Automated Control Symposium on System Identification}, 2021, pp. 481--486.

\bibitem{valasek2003observation}
J.~Valasek and W.~Chen, ``Observer/{Kalman} {filter} {identification} for
  {online} {system} {identification} of {aircraft},'' \emph{AIAA Journal of
  Guidance, Control, and Dynamics}, vol.~26, pp. 347--353, 2003.

\bibitem{baum1970maximization}
L.~E. Baum, T.~Petrie, G.~Soules, and N.~Weiss, ``A maximization technique
  occurring in the statistical analysis of probabilistic functions of markov
  chains,'' \emph{The Annals of Mathematical Statistics}, vol.~41, no.~1, pp.
  164 -- 171, 1970.

\bibitem{schreier2012modeling}
M.~Schreier, ``Modeling and adaptive control of a quadrotor,'' in \emph{{IEEE}
  {International} {Conference} on {Mechatronics} and {Automation}}, 2012, pp.
  383--390.

\bibitem{astrom1965optimal}
K.~J. Åström, ``Optimal {Control} of {Markov} {Processes} with {Incomplete}
  {State} {Information} {I},'' \emph{Journal of Mathematical Analysis and
  Applications}, vol.~10, pp. 174--205, 1965.

\bibitem{ott2023sequential}
J.~Ott, E.~Balaban, and M.~J. Kochenderfer, ``Sequential bayesian optimization
  for adaptive informative path planning with multimodal sensing,'' in
  \emph{IEEE International Conference on Robotics and Automation (ICRA)}, 2023.

\bibitem{kurniawati2009sarsop}
H.~Kurniawati, D.~Hsu, and W.~S. Lee, ``{SARSOP}: Efficient point-based {POMDP}
  planning by approximating optimally reachable belief spaces,'' in
  \emph{Robotics: Science and Systems}, 2009.

\bibitem{silver2010monte}
D.~Silver and J.~Veness, ``Monte-{C}arlo planning in large {POMDP}s,'' in
  \emph{Advances in Neural Information Processing Systems (NIPS)}, 2010.

\bibitem{deisenroth2011pilco}
M.~P. Deisenroth and C.~E. Rasmussen, ``{PILCO}: {A} model-based and
  data-efficient approach to policy search,'' in \emph{International Conference
  on Machine Learning (ICML)}, 2011, p. 465–472.

\bibitem{gonzalez2005robust}
O.~R. González and A.~G. Kelkar, ``Robust multivariable control,'' in
  \emph{The Electrical Engineering Handbook}.\hskip 1em plus 0.5em minus
  0.4em\relax Academic Press, 2005, pp. 1037--1047.

\bibitem{egorov2017pomdps}
\BIBentryALTinterwordspacing
M.~Egorov, Z.~N. Sunberg, E.~Balaban, T.~A. Wheeler, J.~K. Gupta, and M.~J.
  Kochenderfer, ``{POMDP}s.jl: a framework for sequential decision making under
  uncertainty,'' \emph{Journal of Machine Learning Research}, vol.~18, no.~26,
  pp. 1--5, 2017. [Online]. Available:
  \url{http://jmlr.org/papers/v18/16-300.html}
\BIBentrySTDinterwordspacing

\bibitem{barto1983neuronlike}
A.~G. Barto, R.~S. Sutton, and C.~W. Anderson, ``Neuronlike adaptive elements
  that can solve difficult learning control problems,'' \emph{{IEEE}
  Transactions on Systems, Man, and Cybernetics}, pp. 834--846, 1983.

\end{thebibliography}

%\newpage
%\input{sections/99_00_appendix.tex}
\end{document}